%% file: ms.tex
\PassOptionsToPackage{usenames,dvipsnames,table,hyperref}{xcolor}
\documentclass[11pt,a4paper]{article}
\usepackage[hyperref]{emnlp2018}
\usepackage{latexsym}

\usepackage[utf8]{inputenc} 
\usepackage[T1]{fontenc}    
\usepackage{hyperref}       
\usepackage{url}            
\usepackage{booktabs}       
\usepackage{amsfonts}       
\usepackage{nicefrac}       
\usepackage{microtype}      
\usepackage{latexsym}
\usepackage{amssymb}
\usepackage{times}
\usepackage{latexsym}
\usepackage{amsmath}
\usepackage{multirow}
\usepackage{amsfonts}
\usepackage{booktabs}
\usepackage{rotating}
\usepackage{subfig}
\usepackage{array}
\usepackage{booktabs}
\usepackage{amsfonts}
\usepackage{enumitem}
\usepackage{makecell}
\usepackage{xcolor}
\usepackage[toc,page]{appendix}
\usepackage{amsmath}
\usepackage{xspace}
\usepackage{mdwlist}
\usepackage{graphics}
\usepackage{color}
\usepackage{rotating}
\usepackage{booktabs}
\usepackage{epsfig}
\usepackage{alltt}
\usepackage{moreverb}
\usepackage{fancyvrb}
\usepackage{enumerate}
\usepackage{colortbl}      
\usepackage{xspace}
\usepackage{relsize}
\usepackage{array}
\usepackage{amsmath}
\usepackage{amsfonts}
\usepackage{txfonts}
\usepackage{caption}
\DeclareCaptionFont{10pt}{\fontsize{10pt}{12pt}\selectfont}
\usepackage{rotating}
\usepackage{latexsym}
\usepackage{multirow}
\usepackage{booktabs}
\usepackage{float}
\usepackage{array,etoolbox}
\usepackage{enumitem}
\usepackage{subfig}
\usepackage{arydshln}
\usepackage{setspace}
\usepackage{float}
\usepackage{mathtools}
\usepackage{bm}
\usepackage{xcolor}
\usepackage[toc,page]{appendix}
\usepackage{soul}
\usepackage[draft]{pgf}
\usepackage{makecell}
\usepackage[normalem]{ulem}
\usepackage{amssymb}
\usepackage{wasysym,textcomp}
\usepackage[toc,page]{appendix}
\usepackage{wrapfig}

\usepackage{tabularx}
\definecolor{vlgray}{gray}{0.95}

\usepackage{etoolbox}
\newtoggle{draft}
\toggletrue{draft}
\iftoggle{draft}{

\newcommand{\scott}[1]{\textcolor{Maroon}{[#1 \textsc{--Scott}]}}
\newcommand{\niket}[1]{\textcolor{Orange}{[#1 \textsc{--Niket}]}}
\newcommand{\bhavana}[1]{\textcolor{Green}{[#1 \textsc{--Bhavana}]}}

\newcommand{\todo}[1]{\textcolor{Magenta}{[\textsc{TODO: }#1 ]}}

}{
\newcommand{\peter}[1]{}
\newcommand{\scott}[1]{}
\newcommand{\niket}[1]{}
\newcommand{\bhavana}[1]{}

\newcommand{\todo}[1]{}

}

\newcommand{\statechange}[1]{\texttt{\textit{#1}}}
\newcommand{\entity}[1]{\texttt{#1}}

\newcommand{\com}[1]{}

\newcommand{\eat}[1]{}
\mathchardef\mhyphen="2D
\newenvironment{ite}{                     
     \parskip 0cm \begin{itemize} \parskip 0cm \parsep 0cm \itemsep 0cm \topsep 0cm}{
        \end{itemize}} 
\newenvironment{enu}{                   
     \parskip 0cm \begin{list}{}{\parsep 0cm \itemsep 0cm \topsep 0cm}}{
       \end{list}} 

\newcommand{\ourmodel}{\textsc{ProStruct}}

\makeatletter
\g@addto@macro\normalsize{%
  \setlength\abovedisplayskip{1pt}
  \setlength\belowdisplayskip{1pt}
  \setlength\abovedisplayshortskip{1pt}
  \setlength\belowdisplayshortskip{1pt}
}
\makeatother

\aclfinalcopy

\title{Reasoning about Actions and State Changes \\ by Injecting Commonsense Knowledge}

\author{
Niket Tandon\textsuperscript{*}, Bhavana Dalvi Mishra\thanks{\textsuperscript{*}Niket Tandon and Bhavana Dalvi Mishra  contributed equally to this work.}, Joel Grus, Wen-tau Yih, Antoine Bosselut, Peter Clark \\ 
Allen Institute for AI, Seattle, WA \\
{\tt $\{$nikett,bhavanad,joelg,scottyih,antoineb,peterc$\}$@allenai.org} \\
}

\date{}

\begin{document}
\maketitle

\begin{abstract}
Comprehending procedural text, e.g., a paragraph describing photosynthesis,
requires modeling actions and the state changes they produce, so that
questions about entities at different timepoints can be answered.
Although several recent systems have shown impressive progress in this
task, their predictions can be globally inconsistent or highly improbable.
In this paper, we show how the predicted effects of actions in the context of
a paragraph can be improved in two ways:
(1) by incorporating global, commonsense constraints (e.g., a non-existent
entity cannot be destroyed), and (2) by biasing reading with preferences from
large-scale corpora (e.g., trees rarely move). Unlike earlier methods,
we treat the problem as a neural structured prediction task, allowing
hard and soft constraints to steer the model away from unlikely predictions.
We show that the new model significantly outperforms earlier systems
on a benchmark dataset for procedural text comprehension (+8\% relative gain), 
and that it also avoids some of the nonsensical predictions that
earlier systems make. 
\end{abstract}

\eat{
Our goal is to demonstrate process paragraph comprehension by
predicting, tracking, and answering questions about how entities change during a process (e.g., photosynthesis).
Although several recent systems have shown impressive progress in this
task, their predictions can be globally inconsistent or highly improbable.
In this paper, we show how to substantially improve 
process paragraph comprehension in two ways: (1) by incorporating global, commonsense constraints
(e.g., a non-existent entity cannot be destroyed), and 
(2) by biasing reading with preferences from large-scale corpora
(e.g., trees rarely move). Unlike earlier methods,
we treat process comprehension as a structured prediction task, allowing us to
apply both hard and soft constraints during reading.
We show that the new model significantly outperforms earlier systems
on a benchmark dataset for process comprehension, and it also avoids some
of the nonsensical predictions that earlier systems make.
We will release our model code to the community. 
}

\input{intro}
\input{related}

\input{problem}

\input{model}
\input{knowledge}

\input{evaluation}
\input{results}

\input{conclusion}


\bibliography{references}
\bibliographystyle{acl_natbib_nourl}

\end{document}

%% file: intro.tex
\section{Introduction}
Procedural text is ubiquitous (e.g., scientific protocols, news articles, how-to guides, recipes),
but is challenging to comprehend because of the dynamic nature of the world being described.
Comprehending such text requires a model of the actions described in the text and the state changes
they produce, so that questions about the states of entities at different timepoints can be answered \cite{npn}.

\begin{figure}[t]
{\bf Procedural Text:} \vspace{1mm} \\
\fbox{%
    \parbox{0.44\textwidth}{%
    \underline{\bf How hydroelectric electricity is generated:} \\
    1 Water flows downwards thanks to gravity. \\
    2 The moving water spins the turbines in the power plant. \\
    3 The turbines turn the generators. \\
    4 The generators spin, and produce electricity.
        }} \vspace{2mm} \\
  {\bf Prior Neural Model's Predictions:}
\begin{center}
{\includegraphics[width=0.86\columnwidth]{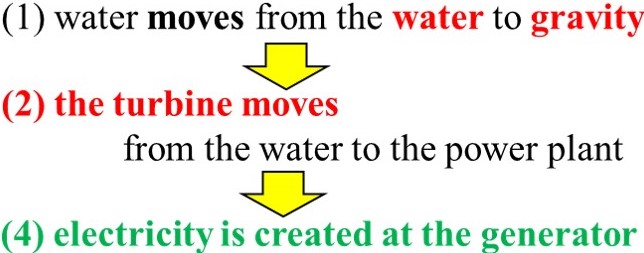}}
\end{center}
\caption{Poor predictions (in red) made by a prior neural model (ProGlobal)
applied to an (abbreviated) paragraph from the ProPara
dataset. ProGlobal predicts entity locations at each sentence,
but the implied movements violate commonsense constraints
(e.g., an object cannot move from itself (1)) and
corpus-based preferences (e.g., it is rare to see
turbines move (2)).
}
\label{example}
\end{figure}

Despite these challenges, substantial progress has been made recently in this task.
Recent work --
such as EntNet \cite{Henaff2016TrackingTW}, 
QRN \cite{Seo2017QueryReductionNF},  ProLocal/ProGlobal \cite{propara-naacl18}, 
and NPN \cite{npn} --
has focused on learning to predict individual entity states 
at various points in the text, thereby approximating the underlying dynamics of the world.
However, while these models can learn to make local predictions with fair accuracy, 
their results are often globally unlikely or inconsistent. For example, in Figure~\ref{example},
the neural ProGlobal model from \citet{propara-naacl18} learns to predict the impossible action of
an object moving from itself (1), and the unlikely action
of a turbine changing location (2). 
We observe similar mistakes in other neural models, 
indicating that these models have little 
notion of global consistency. Unsurprisingly, mistakes in  local predictions compound 
as the process becomes longer, further reducing the plausibility of the overall result.

To address this challenge, we treat process comprehension as a 
{\it structured prediction task} and apply hard and soft constraints
during reading. During training, our model, called \ourmodel, learns to search for the most likely
action sequence that is consistent with global constraints (e.g., entities cannot 
be destroyed after they have already been destroyed) and priors from background knowledge 
(e.g., turbines rarely change location).  
The model is trained end-to-end, with gradients 
backpropagating through the search path. We find that this approach significantly outperforms existing 
approaches on a benchmark dataset for process comprehension, 
mainly by avoiding the nonsensical predictions that earlier systems make.

Our contributions are twofold. First, we reformulate procedural text
comprehension in a novel way: as a (neural) structured prediction task. This lets hard and 
soft constraints  steer the model away from unlikely and nonsensical
predictions. Second, we present a novel, end-to-end model that 
integrates these constraints and achieves state-of-the-art performance 
on an existing process comprehension dataset \cite{propara-naacl18}.

%% file: related.tex
\section{Related Work}
\label{sec:related}
Our work builds off a recent body of work that focuses on using neural networks to explicitly track the states of entities while reading long texts. These works have focused on answering simple commonsense questions \cite{Henaff2016TrackingTW}, tracking entity states in scientific processes \cite{propara-naacl18, propara-arxiv}, tracking ingredients in cooking recipes \cite{npn}, and tracking the emotional reactions and motivations of characters in simple stories \cite{Rashkin2018psychology}. Our work extends these methods and addresses their most common issues by using background knowledge about entities to prune the set of state changes they can experience as the model reads new text.

Prior to these neural approaches, some earlier systems for process comprehension did make use of world knowledge, and motivated this work.
Like us, the system ProRead \cite{berant2014modeling,Scaria2013LearningBP} also treated process comprehension as structure prediction,
using an Integer Linear Programming (ILP) formalism to enforce global constraints (e.g., if the result of event1 is the agent of
event2, then event1 must enable event2). Similarly, \citet{kiddon2015mise} used corpus-based priors to guide
extraction of an ``action graph'' from recipes. Our work here can viewed as incorporating these approaches
within the neural paradigm.

Neural methods for structure prediction have been used extensively in other areas of NLP,
and we leverage these methods here. In particular we use a neural encoder-decoder architecture
with beam search decoding, representative of several current state-of-the-art systems
\cite{Bahdanau2014NeuralMT,Wiseman2016SequencetoSequenceLA,Vinyals2015GrammarAA}.
As our model's only supervision signal comes from the final prediction (of state changes), our work is similar to
previous work in semantic parsing that extracts structured outputs from text with no intermediate supervision \cite{Krishnamurthy2017NeuralSP}.

State tracking also appears in other areas of AI, such as dialog. A typical dialog state tracking task (e.g., the DSTC competitions) involves gradually uncovering the user's state (e.g., their constraints, preferences, and goals for booking a restaurant), until an answer can be provided. Although this context is somewhat different (the primary goal being state discovery from weak dialog evidence), state tracking techniques originally designed for procedural text have been successfully applied in this context also \cite{Liu2017DialogST}.

Finally, our model learns to search over the best candidate structures using hard constraints and soft KB priors. Previous work in Neural Machine Translation (NMT) has used sets of example-specific lexical constraints in beam search decoding to only produce translations that satisfy every constraint in the set \cite{Hokamp2017constraint}.  In contrast, our work uses a set of global example-free constraints to prune the set of possible paths the search algorithm can explore.  Simultaneously, a recent body of work has explored encoding soft constraints as an additional loss term in the training objective for dialogue \cite{Wen2015SemanticallyCL}, machine translation \cite{Tu2016CoveragebasedNM}, and recipe generation \cite{kiddon2016globally}. Our work instead uses soft constraints to re-rank candidate structures and is not directly encoded in the loss function.

%% file: problem.tex
\begin{figure}[t]
\begin{center}
{\includegraphics[width=\columnwidth]{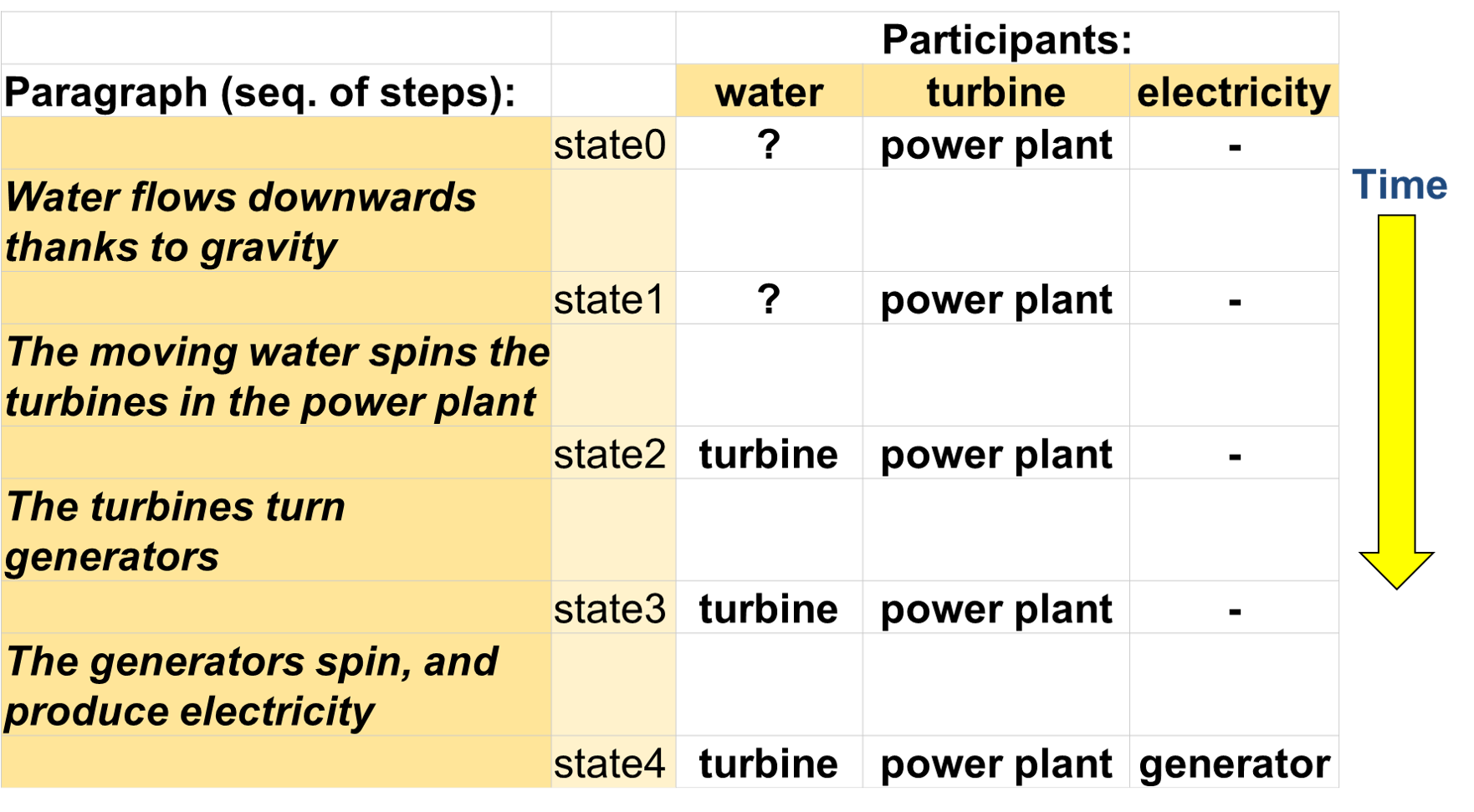}} 
\end{center}
\caption{How the (simplified) paragraph in Figure~\ref{example} is annotated in ProPara. Each
  filled row shows the location of entities between each step (``?'' denotes ``unknown'',
  ``-'' denotes ``does not exist''). For example, in the last line (state4), the
  water is at the turbine.}
\label{propara-example}
\end{figure}

\section{Problem Definition}
\label{sec:problem-definition}
\noindent

We first define the general task that we are addressing, before presenting our approach. 

\subsection{General Formulation}

We define the task as follows. {\bf Given:}
\begin{ite}
\item A {\bf paragraph of procedural text $S$} = an ordered set of sentences \{$s_{1},...,s_{T}$\} describing
  a sequence of {\bf actions}\footnote{
	  We use a broad definition
	  of action to mean any event that changes the state of the world (including non-volitional
	  events such as roots absorbing water).}
  about a given {\bf topic} (a word or phrase).
\item A {\bf set of entities $E = \{e_{j}\}$} representing the entities mentioned in the procedure or process.
  Each entity $e_{j}$ is denoted by the set of its mentions in the paragraph, e.g., \{\entity{leaf, leaves}\}
\item A {\bf set of properties $P =\{p_{k}\}$} of entities to be tracked (e.g., location, existence) 
\end{ite}
{\bf predict:}
\begin{ite}
\item The {\bf state} of each entity $e_{j}$ after each sentence $s_{k}$, where an entity's state
  is the values of all its properties $\{p_{k}\}$.  For example, in Figure~\ref{propara-example}, the state of
  the water after step 2 is \{location(water) = turbine; exists(water) = true\}.
\end{ite}

\noindent This task definition covers the tasks used in earlier procedural text comprehension datasets.
In bAbI tasks 1-3, a single propert (location) was tracked for a single entity throughout a paragraph \cite{weston2015towards}. In the state tracking task of \citet{npn}, six properties (temperature, shape, etc.)
were tracked for each ingredient in the recipe.

\subsection{Data}

In our work, we use the ProPara dataset \cite{propara-naacl18} for both illustration and evalution.
ProPara contains 488 paragraphs (3100 sentences)
of a particular genre of procedural text, namely science processes
(e.g., how hydroelectricity is generated). The dataset
tracks two entity properties, existence and location, for all entities involved
in each process, resulting in 81,000 annotations in the dataset. Figure~\ref{propara-example}
gives a (simplified) example of the data, visualized as an (entity x sentence) grid, where
each column tracks a different entity (time progressing vertically downwards), and each row
denotes the entities' state (existence and location) after each sentence.
To evaluate the predictions, a set of templated questions
whose answers can be computed from the predictions is
posed (e.g., ``What was destroyed, when and where?'').

%% file: model.tex
\section{Model}

We now describe our model, called \ourmodel.

\subsection{Overview}

\eat{
\ourmodel~uses a novel architecture:
Given a sentence (in context) describing an action, we represent the action with N embeddings, one for each of the N entities in the text.
Each embedding represents the action's effects on that entity, allowing us to model different effects on different
entities by the same action. For example, a conversion action may simultaneously destroy one entity and create another.
This contrasts with NPN \cite{npn}, where an action is encoded with a single embedding, and applied to all entities selected as
relevant (i.e., affected). It also contrasts with memory network architectures (e.g., EntNet \cite{Henaff2016TrackingTW}) and ProGlobal \cite{propara-naacl18}, where there is no
explicit action encoding, rather an entity's state is predicted directly from the text and the entity's previous state.

In \ourmodel, these N embeddings are then decoded to predict
the state {\it changes} that occur at
}

We approach the task by predicting the state {\it changes} that occur at
each step of the text, using a vocabulary (size $K$) of the possible
state change types that can occur given the domain and properties being modeled. 
For example, for the ProPara dataset, we model $K=4$ types of
state change: \statechange{move, create, destroy,} and \statechange{none}.
\statechange{move} changes an entity's location from one place to another,
\statechange{create} from non-existence to a location, and
\statechange{destroy} from a location to non-existence.
State changes can be parameterized by text spans in the paragraph,
e.g., \statechange{move} takes a before and after location parameter.
If a parameterized state change is predicted, then the model also must
predict its parameter values from the paragraph. 

Previous models for process comprehension make a sequence of local
predictions about the entities' states, one sentence at a time,
maintaining a (typically neural) state at each sentence. However, none
have the ability to reverse earlier predictions should an inconsistency arise
later in the sequence. \ourmodel~overcomes this limitation by reformulating
the task as structured prediction. To do this, it uses a neural
encoder-decoder from the semantic parsing literature \cite{Krishnamurthy2017NeuralSP,Yin2017ASN}
combined with a search procedure that integrates soft and hard constraints for finding
the best candidate structure.

\begin{figure}[t]
\begin{center}
{\includegraphics[width=1.05\columnwidth]{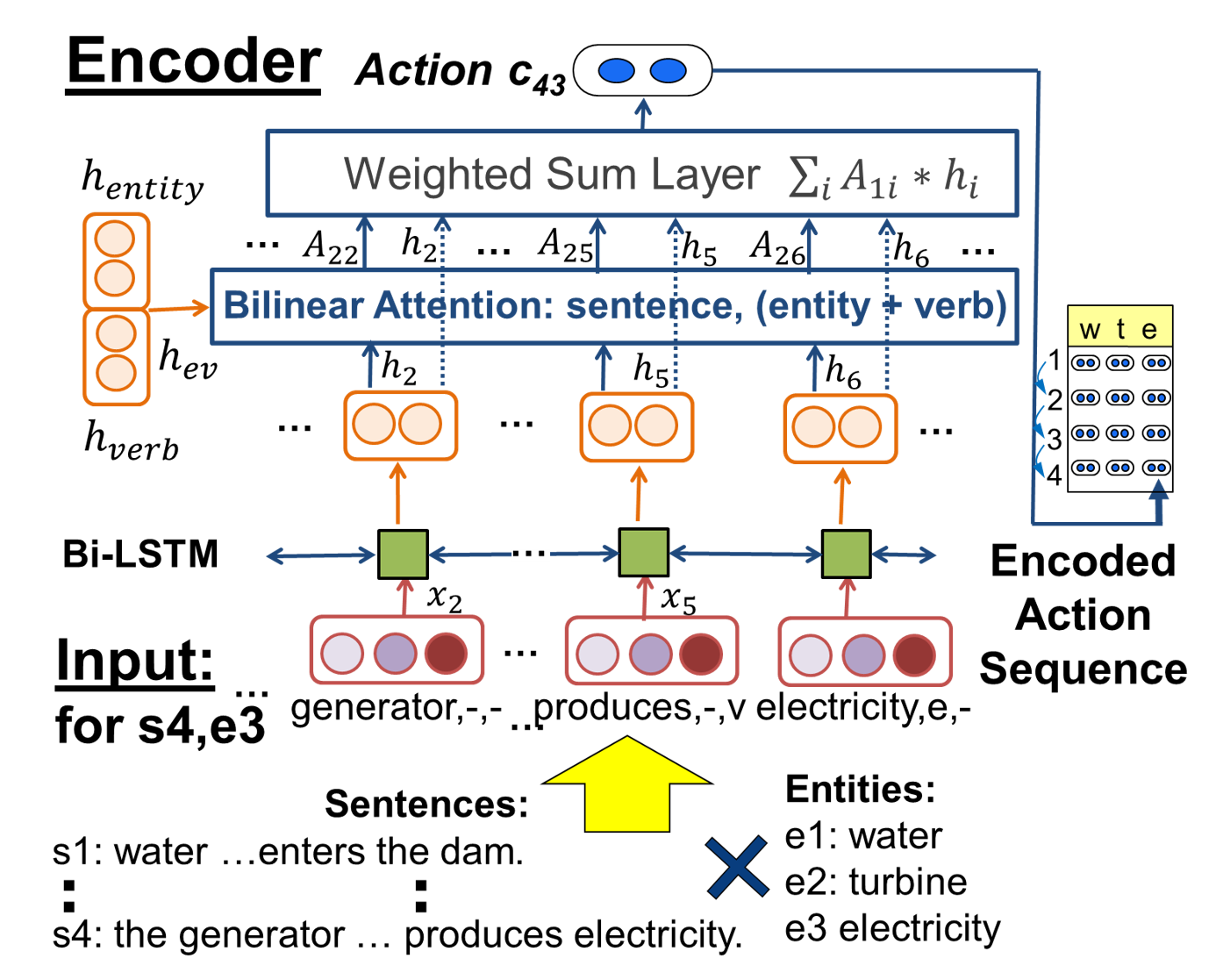}}
\end{center}
\caption{The encoder, illustrated for the ProPara domain with the paragraph from Figure~\ref{example}.
  During encoding, \ourmodel~creates an action embedding $c_{tj}$ representing the action at step $t$ on entity $e_k$,
  for all entities at all steps.
  The overall action sequence (right-hand box) is the collection of these embeddings,
for each entity (listed horizontally) and each step (listed  vertically downwards).}
  \label{encoder}
\end{figure}

For each sentence and entity, the encoder first uses a bidirectional LSTM to encode the sentence
and indicator variables identifying which entity is currently being considered (Figure~\ref{encoder}).
It then produces a (distributed) representation of the action that
the sentence describes as being applied to that entity. During decoding, the model decodes each action embedding into a distribution
over possible state changes that might result, then 
performs a search over the space of possible state change sequences.
Each node in the space is a partial sequence of state changes, and each edge is a prediction of the
next state changes to add to the sequence (Figure~\ref{decoder}).

During training, the model only follows the path along the gold sequence,
and optimizes a loss function that drives up the likelihood of predictions
along that path (thus driving down the probabilities for alternative,
incorrect paths).
At test time, the
model does not have access to the gold path, and instead performs a beam search
of the space to find the best candidate sequence.

Most importantly, by mapping the state change prediction problem to structured prediction, 
we can perform a search over the set of candidate paths that allows us to introduce hard and soft constraints that capture
commonsense knowledge. Hard constraints are used to prune the search space
(Equation~\ref{eq:hard-constraints} later), and soft constraints bias
the search away from unlikely state changes via an additional
term in the scoring function (Equations~\ref{eq:soft-constraints} and~\ref{eq:kb}).

\subsection{Encoder}

The encoder operates over every $(s_t,e_j) \in S \times E$ pair to create an encoded representation $c_{tj}$ of the action described
by sentence $s_{t}$, as applied to entity $e_{j}$. 
In other words, we can consider the overall action to be represented by $\vert E \vert$ embeddings, one for each of the entities in $E$, encoding the action's effects on each.
This novel feature allows us to model different effects on different entities by the same action. For example,
a conversion action may simultaneously destroy one entity and create another.
Figure \ref{encoder} shows the encoder operating on
$s_4$: ``The generator spins, and produces electricity'' and $e_3$: \entity{electricity} from Figure~\ref{example}.

Without loss of generality, we define an arbitrary sentence in $S$ as $s_t = \{w_0, ..., w_I \}$. Each word $w_{i}$ in the input sentence is encoded as a vector $x_{i}=[v_w:v_e:v_v]$, which is the concatenation of a pre-trained word embedding $v_w$ for $w_i$, an indicator variable $v_e$ for whether $w_{i}$ is a reference to the specified entity $e_j$, and an indicator variable $v_v$ for whether $w_{i}$ is a verb. We use GloVe vectors as pre-trained embeddings \cite{pennington2014glove} and a POS tagger to extract verbs \cite{spacy-tagger}. 

Then, a BiLSTM is used to encode the word representations extracted above, yielding a contextualized vector $h_i$ 
for each embedded word $x_{i}$ that is the concatenated output of the backward and forward hidden states produced by the BiLSTM for word $w_i$. 
An attention over the contextualized embeddings $h_i$ is performed to predict a distribution of weights over the sentence:

\begin{align}
    a_i &= h_i * B * h_{ev} + b \\
    c_{tj} &= \sum_{i=1}^I a_{i} * h_i
\end{align}

\noindent where $a_i$ is the attention weight for each contextualized embedding, $c_{tj}$ is the vector encoding the action for the sentence-entity pair ($s_t,e_j$), $B$ and $b$ are learned parameters, and $h_{ev}$ is the concatenation of the contextual embeddings of the hidden states where the entity $h_e$ and verb $h_v$ are mentioned:

\begin{equation}
    h_{ev} = [\mu(\{h_i : x_i[v_e] = 1 \}); \mu(\{h_i : x_i[v_v] = 1\}]
\end{equation}

\noindent where $\mu$ is an average function, and $x_i[v_e]$ and $x_i[v_v]$ correspond to the entity indicator and verb indicator variables defined above for any word $w_i$, respectively.
The output vector $c_{tj}$ encodes the action at step $s_t$ on entity $e_j$. This vector is computed for all steps and entities, populating a grid of the actions on each entity at each step (Figure~\ref{encoder}).

\subsection{Decoder}

\begin{figure}[t]
\begin{center}
{\includegraphics[width=1.05\columnwidth]{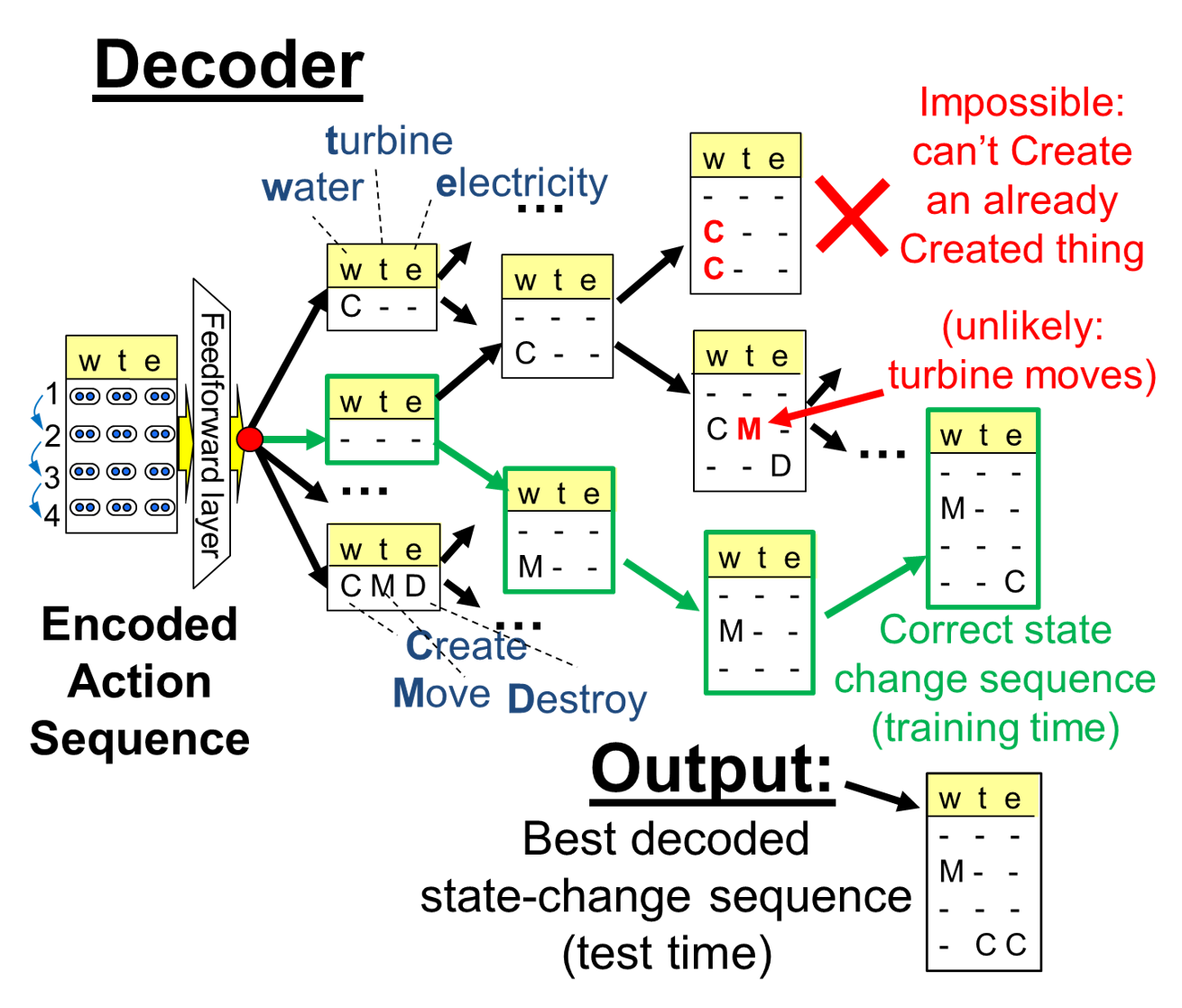}}
\end{center}
\caption{The decoder, illustrated for the ProPara domain. Each action embedding $c_{tj}$ is first
  passed through a feedforward layer to generate a distribution over the (here $K=4$) possible
  state changes that could result, for each entity (listed horizontally) at each step (listed vertically downwards). The decoder then explores the space
   of state-change sequences, using these distributions to guide the search.
 During end-to-end training, \ourmodel~follows the correct (green) path, and backpropagates to drive
 up probabilities along this path.
 During testing, the system performs a beam search to find the most globally plausible sequence.}

  \label{decoder}
\end{figure}

To decode the action vectors $c_{tj}$ into their resulting state changes they imply, 
each is passed through a feedforward layer to generate $logit(\pi^{j}_{t})$, a set of logistic
activations over the $K$ possible state changes $\pi_t^j$ for entity $e_j$ in sentence $s_t$.
(For ProPara, there are $K=4$ possible state changes: \statechange{move}, \statechange{create}, \statechange{destroy}, \statechange{none}).
These logits denote how likely each state change $\pi^j_{t}$ is for entity $e_j$ at sentence $s_t$.
The decoder then explores the search space of possible state change sequences
for the \textbf{whole} paragraph (Figure~\ref{decoder}), using these likelihoods to score each visited sequence (Equation~\ref{eq:kb}).

Let $\pi_t$ be the set of state changes for {\it all} entities at time $t$, i.e., $\pi_t = \{\pi^j_t\}_{j=1..\vert E \vert}$,
and let $\Pi_t$ be the sequence of state changes from time 1 to $t$, i.e., $\Pi_t = [\pi_1,...,\pi_t]$.
Each node in the search space is a $\Pi_t$, and each edge adds a $\pi_{t+1}$ to it so that it becomes $\Pi_{t+1}$:
\begin{equation*}
  \Pi_t \xrightarrow[]{\pi_{t+1}} \Pi_{t+1}
  \end{equation*}
Given there are $K$ possible values for $\pi^{j}_{t}$, the number of possible configurations for $\pi_{t}$ at time $t$ (i.e., the branching factor during search) is exponential: $K^{\vert E\vert}$, where $\vert E \vert$ is the number of entities in the paragraph.

To explore this exponential number of paths, after every sentence $s_t$, 
we prune branches $\Pi_t \rightarrow \Pi_{t+1}$ where $\Pi_{t+1}$ is impossible according to background knowledge (described in Section \ref{sec:hard-constraints}). We define the boolean function over state change sequences:
\begin{align} \label{eq:hard-constraints}
  \text{allowable}(\Pi) & = 1~\text{if hard constraints satisfied} \nonumber \\
  & = 0~\text{otherwise} 
\end{align}
and prune paths $\Pi_{t+1}$ where $allowable(\Pi_{t+1}) = 0$. 
For example for ProPara, a state transition such as 
$\statechange{DESTROY} \rightarrow \statechange{MOVE}$
is not allowed because a hard constraint prohibits non-existent entities from
being moved (Section~\ref{sec:hard-constraints}).

While hard constraints remove impossible state change predictions, there may also be other state changes
that are implausible with respect to background knowledge.
For example, commonsense dictates that it is unlikely (but not impossible) for plants to be destroyed during photosynthesis. 
Accordingly, our inference procedure should discourage (but not prohibit) predicting plant destruction when reading about photosynthesis.
To discourage unlikely state changes, we make use of soft constraints that estimate the likelihood of a particular
state change associated with an entity, denoted as:
\begin{equation} \label{eq:soft-constraints}
  P(\pi^j | e_j,  topic)
\end{equation}
In Section~\ref{sec:soft-constraints}, we describe how these likelihoods can be estimated from large-scale corpora.
We add this bias as an additional term (the second term below) when scoring the addition of $\pi_{t+1}$ to the sequence so far $\Pi_t$:
\begin{align}
\label{eq:kb}
\phi'(\pi_{t+1}) &= \sum_{j=1}^{\vert E \vert} \Big(  \lambda \text{ } logit(\pi_{t+1}^j)  \nonumber \\
                            &\qquad {} + (1 - \lambda) \text{ }  \log P(\pi_{t+1}^j | e_j,  topic) \Big) 
\end{align}
\noindent where $\lambda$ is a learned parameter controlling the degree of bias.

During search, when making a transition along a path from $\Pi_{t}$ to a valid $\Pi_{t+1}$,
$\Pi_{t+1}$ is scored by accumulating normalized scores along the path:

\begin{equation} \label{normalized-scores}
\phi(\Pi_{t+1}) = \phi(\Pi_{t}) + \frac{\phi'(\pi_{t+1})}{\sum_{\pi'_{t+1} \in \Pi_{t+1}}{\phi'(\pi'_{t+1})}} 
\end{equation}
\noindent
Continuing state transitions in this manner, when we reach the finished state (i.e., last sentence), our objective is to maximize the score of the state changes produced when reading each sentence. During training, we only materialize a valid node when $\Pi_t \in \Pi_t^*$ where $\Pi_t^*$ is the set of nodes along the gold path.

\eat{ This is the specific version:
  We use this constrained decoding only for predicting the state change sequence. Conditioned on the state change sequence produced, we also predict the location spans that describe \emph{where} each entity is at any point in time.  
Before and after location spans are predicted from the sentence whenever any entity undergoes a \statechange{MOVE} state change.
We use standard span prediction layers (inspired by BiDAF, \citet{Seo2016BidirectionalAF}) on top of the encoded input for location prediction.}
 
We use this constrained decoding to predict the state change sequence.
For state changes that take additional parameters, e.g., in the ProPara
model a \statechange{move} is parameterized by the before
and after locations, we also predict those parameter values
during decoding. This is done using standard span prediction layers
(inspired by BiDAF, \citet{Seo2016BidirectionalAF}) on top of
the encoded input. 

The model is trained to minimize the joint loss of predicting the correct state changes and correct state change parameters for every sentence in the paragraph:

\begin{equation}
\label{eq:loss}
\mathcal{L} =  - \sum_{t=1}^T \Big(\log P(\pi_t)  + \sum_{j=1}^{\vert E \vert} \sum_{p \in \text{param}(\pi^j_t)} \log P(v_{pjt} \vert \pi^j_t) \Big)
\end{equation}
\noindent
where param($\pi^j_t$) are the parameters of state change $\pi^j_t$, and $v_{pjt}$ are the values of those parameters.
For example, \statechange{move} is parameterized by before/after locations, and the 2nd loss term refers to
the predicted values of those locations.

At test time, instead of following the gold state change path, we use beam search. After reading any sentence, we explore the top-$k$ states sorted by the score $\phi'(\pi_{t})$ that satisfy hard constraints. 
This way, we predict a sequence of state changes that have maximum score while being sensible w.r.t. hard constraints.

%% file: knowledge.tex
\section{Incorporating Commonsense Knowledge}
\label{sec:constraints}

By formulating procedural text comprehension as a structured prediction task,
we can introduce commonsense knowledge as hard and soft constraints into
the model, allowing nonsensical and unlikely predictions to be avoided,
and allowing the system to recover from early mistakes.

\subsection{Hard Constraints}
\label{sec:hard-constraints}

Hard constraints are introduced by defining the (boolean) function over a candidate sequence of state changes:
\begin{equation*}
 \text{allowable}(\Pi)
\end{equation*}
\noindent
used in Equation~\ref{eq:hard-constraints}.

While this function can be defined in any way, for the ProPara application we use six constraints.
The first three below are based on basic ``laws of physics'' or commonsense (CS) and are universally applicable:
\begin{ite}
    \item[CS-1:] An entity must \textbf{exist} before it can be \textbf{moved} or \textbf{destroyed}
    \item[CS-2:] An entity cannot be \textbf{created} if it already \textbf{exists}
    \item[CS-3:] An entity cannot change until it is mentioned in the paragraph
\end{ite}

The next three constraints are observed in the training data: 
\begin{ite}
    \item[D-1:] Maximum number of toggles for an entity between Exists and not Exist $\leq f_{max\_toggles}$
    \item[D-2:] Max fraction of entities that are changed per sentence $\leq f_{entities\_per\_sentence}$
    \item[D-3:] Max fraction of sentences in which an entity changes  $\leq f_{sentences\_per\_entity}$
\end{ite}
The thresholds used in D-1, D-2 and D-3 are hyperparameters that can be tuned on the dev set.

\subsection{Soft Constraints}
\label{sec:soft-constraints}

Soft constraints are introduced by defining the prior probabilities used in Equation~\ref{eq:kb}:
\begin{equation*}
  P(\pi^j | e_j,topic)
\end{equation*}
that entity $e_j$ undergoes state change $\pi^j$ in a sentence of text about $topic$.
These probabilities are used to re-rank the candidate event sequences during decoding (see Equation \ref{eq:kb}).

While any method can be used to estimate these probabilities,
we describe our corpus-based approach here. Although it was designed for ProPara, it generalizes easily to other
domains, and is itself a contribution of this work. For a given state change $\pi^{j}$, entity $e_{j}$, and topic, we first gather
a corpus of Web sentences mentioning that topic (using Bing search APIs), then count the number of times $x$ 
that the entity is described as undergoing that state change (e.g., that \entity{water} is said to \statechange{MOVE}).
To determine this frequency, we first convert the sentences into a set of SRL frames (verb + role-argument pairs) using an off-the-shelf
SRL labeler. We then use an existing rulebase, derived from VerbNet, that contains rules that map SRL frames to state changes,
e.g., \entity{e1/ARG0 ``absorbs''/VERB e2/ARG1} $\implies$ \entity{e2 MOVES} \cite{propara-arxiv}.
Although the rules and SRL labels are incomplete and noisy, redundancy in the corpus provides
some robustness when estimating the frequency $x$. Finally, the observed frequency $x$ is converted to a likelihood
using a logistic transformation:
\begin{equation}
    P(\pi^j | e_j,  topic) = \frac{1}{1 + exp^{-(x-x_0)}}
\end{equation}
where, $x_0$ is a hyperparameter tuned on the dev set.

\subsection{Commonsense Constraints for New Domains}

The commonsense constraints we have used for ProPara are general, covering the large variety of 
topics contain in ProPara (e.g., electricity, photosynthesis, earthquakes).
However, if one wants to apply ProStruct to other genres of procedural
text (e.g., fictional text, newswire articles), or broaden the
state change vocabulary, different commonsense constraints may be needed.
Note that our model architecture itself is agnostic to the source and
quantity of hard and soft constraints. 
For example, one might leverage commonsense rules from existing ontologies
such as SUMO \cite{Niles2001TowardsAS} or Cyc \cite{lenat1985cyc} to identify
new hard constraints; and our corpus-based method could be extended to
cover new state change types should the state change vocabulary be extended.


%% file: evaluation.tex
\section{Evaluation}

We evaluate our model using the ProPara dataset, and compare against several strong baselines published with the original dataset \cite{propara-naacl18}. 

\subsection{Evaluation setup}  

Given a paragraph and set of entities as input, the
task is to answer four templated questions, whose answers are deterministically computed from the state change sequence:
\begin{enu}
\item[Q1.] What are the inputs to the process?
\item[Q2.] What are the outputs of the process?
\item[Q3.] What conversions occur, when and where?
\item[Q4.] What movements occur, when and where?
\end{enu}
Inputs are defined as entities that existed at the start of the process, but not at the end. Outputs are entities that did not exist at the start, but did at the end. A conversion is when some entities are destroyed and others created. Finally, a movement is an event where an entity changes location.

For each process, as every question can have multiple answers, we compute a separate F1 score for each question by comparing the gold and predicted answers. For Q1 and Q2, this is straightforward as answers are atomic (i.e., individual names of entities). For Q3, as each answer is a 4-tuple ({\it convert-from, convert-to, location, sentence-id}), some answers may only be partially correct. To score partial correctness, we pair gold and predicted answers by requiring the {\it sentence-id} in each to be the same, and then score each pair by the Hamming distance of their tuples.
For Q4, each answer is also a 4-tuple ({\it entity, from-location, to-location, sentence-id}), and the same procedure is applied. The four F1 scores are then macro-averaged. The total number of 
items to predict in the train/dev/test partitions is 7043/913/1095.

\subsection{Baselines}

We compare results using the following process comprehension models:

\noindent
{\bf Recurrent Entity Networks (EntNet)}~\cite{Henaff2016TrackingTW}
are a state-of-the-art model for the bAbI tasks~\cite{weston2015towards}.
The model uses a dynamic memory to maintain a representation
of the world state as sentences are read, with a gated update at each step. These states are decoded to answer questions after each sentence is read.  

\noindent
{\bf Query Reduction Networks (QRN)}~\cite{Seo2017QueryReductionNF} perform a gated propagation of their hidden state across each time-step. Given a question, the hidden state is used to modify the query to keep pointing to the answer at each step.

\noindent
{\bf ProLocal}~\cite{propara-naacl18} predicts the state changes
described in individual sentences, and then uses commonsense rules of
inertia to propagate state values forwards and backwards in time.

\noindent
{\bf ProGlobal}~\cite{propara-naacl18} predicts states of an entity across all time steps. It considers the entire paragraph while predicting states for an entity, and learns to predict location spans at time-step $t+1$ based on location span predictions at $t$.

%% file: results.tex
\section{Results}
\subsection{Comparison with Baselines}

We compare our model (which make use of world knowledge) with the four baseline systems on the ProPara dataset.  All models were trained on the training partition, and the best model picked based on prediction accuracy on the dev partition. Table~\ref{table:F1-QA-task} shows the  precision, recall, and F1 for all models on the the test partition. \ourmodel~significantly outperforms the baselines, suggesting that world knowledge helps \ourmodel~avoid spurious predictions. This hypothesis is supported by the fact that the ProGlobal model has the highest recall and worst precision, indicating that it is over-generating state change predictions. Conversely, the ProLocal model has the highest precision, but its recall is much lower, likely because it makes predictions for individual sentences, and thus has no access to information in surrounding sentences that may suggest a state change is occurring.

\begin{table}[tbh]
\centering
\boldmath
 \begin{tabular}{|l|ccc|} \hline
	      & Precision & Recall & F1\\ \hline
ProLocal  & 77.4 & 22.9 & 35.3 \\
QRN       & 55.5 & 31.3 & 40.0 \\
EntNet    & 50.2 & 33.5 & 40.2 \\
ProGlobal & 46.7 & 52.4 & 49.4 \\
\hline 
{\bf \ourmodel} & 74.2 & 42.1 & \textbf{53.7} \\ \hline
 \end{tabular}
\caption{Results on the prediction task (test set). }
\label{table:F1-QA-task}
\vspace{-2mm}
\end{table}

We also examined the role of the constraint rules (both hard and soft) on efficiency.  With all rules disabled, the training does not complete even one epoch in more than three hours. Because the number of valid states is exponential in the number of entities, the training is particularly slow on paragraphs with many entities. In contrast, with all rules enabled, training takes less than 10 minutes per epoch. This illustrates that the constraints are not only contributing to the model scores, but also helping make the search efficient.

\subsection{Ablations and Analysis}
\eat{As a quantitative measure of the impact of the world knowledge constraints, we performed  ablations on the dev set. Table \ref{table:ablations} shows that removing either hard or soft constraints results in a reduced F1, showing that both forms of world knowledge integration help boost performance. We note that hard constraints were only removed at test time because model training time becomes prohibitively large without them, implying that the hard constraints are more important than scores in Table~\ref{table:analysis-kb2} may indicate. }

To explore the impact of world knowledge, we also performed two ablations on the dev set: Removing soft constraints (at both training and test time), and a partial ablation of removing hard constraints at test time only - note that hard constraints cannot be removed during training because model training time becomes prohibitively large without them, thus qualifying this second ablation. Table~\ref{table:analysis-kb2} shows that F1 drops when each type of knowledge is removed, illustrating that they are helping. The smaller drop for hard constraints suggests that they have primarily been incorporated into the network during training due to this ablation being partial.

\begin{table}[t]
\centering
\boldmath
\scalebox{0.85}{%
 \begin{tabular}{|l|ccc|} \hline
	                    & Precision  &  Recall &  F1 \\ \hline
{\bf \ourmodel}                    & 70.4     & 47.8    & 56.9  \\ \hline
\hspace*{2mm} - Soft constraints   & 61.9	  & 47.4    & 53.7 \\ 
\hspace*{2mm} - Hard constraints$^{\dag}$   & 69.6	  & 47.0	& 56.1 \\ \hline
 \end{tabular}
} \\
\small{}
$^{\dag}$ Partial ablation, ablated at test only (training without these is computationally infeasible).
\caption{Ablating world knowledge on the dev set. }
\label{table:ablations}
\vspace{-0.2in}
\end{table}

Qualitatively, we compared dev set examples where the predicted event sequence changed,
comparing predictions made without world knowledge to those made with world knowledge.
For readability, we only show the event type predictions (\statechange{M},\statechange{C},\statechange{D}, and \statechange{N} (shown as "-")) and not their from-location/to-location
arguments. If a prediction changes from \statechange{X} (without knowledge) to \statechange{Y} (with knowledge), we
write this ``\statechange{X} $\rightarrow$ \statechange{Y}''. For cases where the prediction changed, we show
incorrect predictions in red, and correct predictions in green.

We first compare predictions made with and without the \textbf{BK} (corpus-based background knowledge,
the soft constraints). Table~\ref{table:analysis-kb1} shows a paragraph about the process of
nuclear-powered electricity generation, in the problematic prediction of the generator moving (\statechange{M}) was predicted in the second to last sentence.
However, the background knowledge
contains no examples of generators being moved. As a result, it drives the probability mass away from
the move (\statechange{M}) prediction, resulting in a no state change (\statechange{N}) prediction instead.

\begin{table}[b]
\centering
\boldmath
\scalebox{0.70}{%
 \begin{tabular}{|l|ccccc|} \hline
           & \multicolumn{5}{|c|}{Without vs. with BK}\\
	       &  Fuel & Heat & Turbine  &  Generator  &  Elec.
	   \\ \hline
Fuel produces heat.  & D & C & - & - & -  \\
...
Steam spins turbine. & - & -  & - & - & - \\
Generator is turned. & - & -  & - & \textcolor{red}{M} $\rightarrow$ \textcolor{green}{N} & - \\
Makes electricity.   & - & -  & - & - & C \\
\hline
 \end{tabular}
}
\caption{BK improves precision. In a nuclear powered electricity generation scenario, BK drives the probability mass away from the generator movement, as a generator does not generally change location.}
\label{table:analysis-kb1}
\end{table}

Table~\ref{table:analysis-kb2} shows a second example where, without knowledge, no event was predicted for the spark entity. However, BK contains many examples of 
sparks being created (reflecting text about this topic), shifting the probability mass towards this prediction, resulting in the correct C (create).

\eat{Here, we illustrate the strengths and limitations of \ourmodel{} with anecdotal examples. Tables \ref{table:analysis-kb1}, \ref{table:analysis-kb2} depict that soft constraints either remove nonsensical predictions, or increase prediction coverage (also corroborated by ablation numbers in Table \ref{ablations} previously). Similarly, we illustrate in Table \ref{table:analysis-rules} that hard constraints and domain constraints help \ourmodel{} to explore sensible paths. }

\begin{table}[tb]
\centering
\boldmath
\scalebox{0.75}{%
 \begin{tabular}{|l|ccc|} \hline
           & \multicolumn{3}{|c|}{Without vs. with BK }\\
	       &  Fuel & Air & Spark
	   \\ \hline
Fuels burns in the chamber. & D & - & -  \\
The burning fuel creates energy.& - & - & -  \\
The upward motion cause air ... & - & M & -  \\
The piston compresses the air.& - & - & -  \\
A spark ignites the fuel and air ...& - & - & \textcolor{red}{N} $\rightarrow$ \textcolor{green}{C}  \\
... & ... & ... & ...  \\
\hline
 \end{tabular}
}
\caption{BK improves coverage. BK has a strong signal that a spark is usually created in combustion engines, shifting the probability mass towards spark-creation.}
\label{table:analysis-kb2}
\vspace{-3mm}
\end{table}

Finally, Table~\ref{table:analysis-rules} shows an example of a hard constraint preventing a nonsensical prediction (namely, electricity is created after it already exists).

\begin{table}[!h]
\centering
\boldmath
\scalebox{0.66}{%
 \begin{tabular}{|l|ccc|} \hline
          & \multicolumn{3}{|c|}{Without and with constraints}\\
	       &  Electricity & Signals & ... \\ \hline
Electricity enters supply unit.& M & - & -  \\
The supply gives electricity to transistors. & \textcolor{red}{C} $\rightarrow$ \textcolor{green}{D} & - &  \\
... & ... & ... & ...  \\
The energy is used to complete ...& - & - & \\\hline
 \end{tabular}
}
\caption{Hard constraints avoid nonsensical predictions. In this example without $CS\text{-}2$, the electricity is predicted to be created after it already exists (impossible). This mistake
is avoided using the constraints.}
\label{table:analysis-rules}
\vspace{-3mm}
\end{table}

\subsection{Error Analysis}

There are also many cases where incorrect predictions are made. The main causes
are summarized below, and offer opportunities for future work.

{\bf Implicit reference} is a challenge for \ourmodel{}, where an entity affected by an event is not mentioned until a later sentence in the paragraph. For example, in the following ProPara paragraph snippet about combustion engines:
\begin{quote}
\vspace{-1mm}
"...(3) A spark ignites fuel...(4) The pressure pushes the piston down...."
\vspace{-1mm}
\end{quote}
both \entity{spark} and \entity{pressure} are created in sentence 3, even though \entity{pressure} is not mentioned until
the subsequent sentence. Recognizing this type of implicit mention is very hard. It is possible
that BK could help in such situations, particularly if \entity{ignite} were often associated with creating pressure in the context of a combustion engines, but we did not see such examples in practice.

A second challenge is {\bf coreference}, in particular when different entities have similar
names. For example, again for combustion, a snippet looks:
\begin{quote}
\vspace{-1mm}
...(2) the fuel is injected... (6) the spent fuel is ejected. (7) new fuel is injected....
\vspace{-2.5mm}
\end{quote}
Here \entity{fuel} and \entity{spent fuel} are the same entity, while \entity{new fuel} is a different entity.
Correctly tracking these references is challenging (in this case, \ourmodel~misidentifies
(7) as describing an event on the original \entity{fuel/spent fuel}).

A third, related problem is {\bf pronoun resolution}. For example, in:
\begin{quote}
\vspace{-2mm}
The sound continues to bounce off of things and produce echoes until it is totally absorbed or dissipated.
\vspace{-2mm}
\end{quote}
the word \entity{it} confuses \ourmodel{}, and it predicts that the \entity{echo} (rather than the \entity{sound}) is destroyed. We observe several such failure cases.

Finally, we observed {\bf BK retrieval failures} when there was appropriate background knowledge that was expressed in a lexically different way. Consider the example in Table~\ref{table:analysis-kb-lookups} about oil formation. Without BK, the model correctly predicts that \entity{sediment} is destroyed (D). However, BK has few examples of \entity{sediment} being destroyed, and so biases the prediction away from this (correct) choice to an incorrect choice.
Further examination of BK shows that it does in fact have knowledge about this destruction, but that is expressed using the word \entity{deposit} instead (e.g., "deposits break down"). A soft (neural) means of accessing BK would help alleviate this problem.

\begin{table}[!h]
\centering
\boldmath
\vspace{-2mm}
\scalebox{0.73}{%
 \begin{tabular}{|l|ccc|} \hline
           & \multicolumn{3}{|c|}{Without BK vs. with BK }\\
	       &  Algae & Plankton & Sediment
	   \\ \hline
Algae and plankton die.& D & D & -  \\
The dead algae and plankton ... & - & - & -  \\
The sediment breaks down.& - & - & \textcolor{green}{D} $\rightarrow$ \textcolor{red}{M}  \\
\hline
 \end{tabular}
}
\caption{BK lookup limitation: though BK knows
that \entity{deposits} can be destroyed (broken down), it does not equate this with (synonymous) \entity{sediments} being destroyed, hence biases model away from
 correct answer.}
\label{table:analysis-kb-lookups}
\vspace{-4mm}
\end{table}

%% file: conclusion.tex
\eat{
    \todo{Future work}
    \begin{itemize}
    \item Soft KB lookup using embeddings:
      - Distribution over topics and 
      - KB lookup  (sediment/soil)
      - KB topic need not be provided at KB construction time.
      - Maybe we can say that currently recall is low, and this can help improve recall
    \item Adding LSTM across steps
       - Hard constraints to remove nonsensical
       - Learn more likely vs less likely even if sensible
    \item Coref confuses what is being changed
       - So we can add coref resolution in dataset reader to update entity indicators
    \item Moving to real paragraphs 
       - can extract entities and with better RNN learn probable sequences automatically.  (can combine with adding LSTMs)
    \end{itemize}
}
\section{Conclusions}

\eat{
This paper demonstrates that more accurate and more sensible comprehension of procedural text can be achieved by injecting world knowledge. 
We propose \ourmodel, an end-to-end neural model that makes use of world knowledge to efficiently explore an otherwise exponential  search-space to move towards sensible and likely state-change predictions.
\ourmodel{} achieves 10\% F1 relative improvement over state-of-the-art on a process comprehension benchmark. 

As future work, we will apply these insights for comprehension of real procedural paragraphs (not turked) from any domain, requiring exploration along two dimensions:
(i) We can embed the knowledge (soft constraints) in a latent space of topics and entities. This would increase the number of instances on which soft constraints are applied (e.g., destruction of ``entity=deposits'' in ``topic=oil formation'', can give a strong signal about destruction of ``entity=sediment'' in ``topic=gas formation'').
(ii) The domain specific hard constraints can be learned automatically from data, e.g., by adding a seq2seq layer across different steps we can learn which event sequences are more likely compared to others. (e.g., creations are more likely to occur in the later steps as compared to the first step).
}

Answering questions about procedural text remains challenging,
requiring models of actions and the state changes they produce.
Predictions made locally throughout the text may together be
globally inconsistent or improbable. We have shown how
the predicted effects of actions can be improved by treating
the task as a structured prediction problem,
allowing commonsense knowledge to be injected to avoid
an overall inconsistent or improbable set of predictions.
In particular, we have shown how two kinds of knowledge can be
exploited: hard constraints to exclude impossible and nonsensical
state changes, and soft constraints to encourage likely state changes.
The resulting system significantly outperforms previous state-of-the-art systems
on a challenging dataset, and our ablations and analysis suggest that the knowledge is playing
an important role. Our code is available at \url{https://github.com/allenai/propara}.

\section*{Acknowledgements}
We thank Oren Etzioni for his insightful feedback and encouragement for this work.
We are grateful to Paul Allen whose long-term vision continues to inspire our scientific endeavors.

\eat{OLD VERSION
Answering questions about procedural text remains challenging, as the world state
being described is constantly changing, often in implicit ways. Predictions
made locally throughout the text may together be globally inconsistent or improbable.
We have shown how this problem can be alleviated by treating process comprehension as a structured
prediction task, allowing world knowledge to be injected to avoid an overall
inconsistent or improbable set of predictions.
In particular, we have shown how
two kinds of world knowledge can be exploited: hard constraints
to exclude impossible and nonsensical state changes, and soft constraints to encourage likely state changes, using priors determined from an automatically assembled corpus about the process topic.
The resulting system significantly outperforms previous state-of-the-art systems
on a challenging dataset, and our ablations and analysis suggest that the knowledge is playing
an important role. We are making our model code available to the community.
}
